%% file: main_arxiv.tex
\documentclass[11pt]{article}

\usepackage{times}
\usepackage{graphicx} 
\usepackage{subfigure} 

\usepackage{natbib}

\usepackage{algorithm}
\usepackage{algorithmic}

\usepackage{hyperref}

\usepackage{arxiv}

\usepackage{times}
\usepackage{helvet}
\usepackage{courier}

\usepackage{url}
\usepackage{graphicx}

\def\Degsym{D}
\def\reals{\mathbb{R}}
\def\transpos#1{#1^{\top}}
\def\s#1{{\cal #1}}
\def\norme#1{\left\|#1\right\|}

\usepackage{amsthm}
\usepackage{amssymb,amsmath,amsthm,amsfonts}

\newtheorem{proposition}{Proposition}

\theoremstyle{definition}
\newtheorem{definition}{Definition}[section]

\icmltitlerunning{Semantic Word Clusters Using Signed Normalized Graph Cuts}

\begin{document}

\twocolumn[

\icmltitle{Semantic Word Clusters Using Signed Normalized Graph Cuts}

\icmlauthor{Jo\~{a}o Sedoc}{joao@cis.upenn.edu}
\icmladdress{Department of Computer and Information Science, University of Pennsylvania
            Philadelphia, PA 19104 USA}
\icmlauthor{Jean Gallier}{jean@cis.upenn.edu}
\icmladdress{Department of Computer and Information Science, University of Pennsylvania
            Philadelphia, PA 19104 USA}
\icmlauthor{Lyle Ungar}{ungar@cis.upenn.edu}
\icmladdress{Department of Computer and Information Science, University of Pennsylvania
            Philadelphia, PA 19104 USA}
\icmlauthor{Dean Foster}{dean@foster.net}
\icmladdress{Amazon, New York, NY USA}
]

\frenchspacing


\input{sections/abstract}
\input{sections/introduction}

\input{sections/theory}
\input{sections/metrics}
\input{sections/results}
\input{sections/conclusion}

\bibliographystyle{icml2015}
\bibliography{spectral_clustering,ssc}

\end{document}

%% file: sections/abstract.tex
\begin{abstract}
Vector space representations of words capture many aspects of word similarity, but such methods tend to make vector spaces in which antonyms (as well as synonyms) are close to each other.  We present a new signed spectral normalized graph cut algorithm, {\em signed clustering}, that overlays existing thesauri upon distributionally derived vector representations of words, so that antonym relationships between word pairs are represented by negative weights.  Our signed clustering algorithm produces clusters of words which simultaneously capture distributional and synonym relations.  We evaluate these clusters against the SimLex-999 dataset \cite{hill2014simlex} of human judgments of word pair similarities, and also show the benefit of using our clusters to predict the sentiment of a given text. 
\end{abstract}

%% file: sections/introduction.tex
\label{intro}
\section{Introduction}

While vector space models \citep{turney2010frequency} such as Eigenwords, Glove, or word2vec capture relatedness, they do not adequately encode synonymy and similarity \citep{mohammad2013computing,scheible2013uncovering}. 
Our goal was to create clusters of synonyms or semantically-equivalent words 
and linguistically-motivated unified constructs. 
We innovated a novel theory and method that extends multiclass normalized cuts 
(K-cluster) to signed graphs \citep{gallier2015spectral}, which
allows the incorporation of semi-supervised 
information. Negative edges serve as repellent or opposite relationships between nodes. 

In distributional vector representations opposite relations
are not fully captured. Take, for example, words such as ``great'' and ``awful'', which can appear with similar frequency in the same sentence structure: ``today is a great day'' and ``today is an awful day''. 
Word embeddings, which are successful in a wide array of NLP tasks, fail to capture this antonymy because they follow the {\it distributional hypothesis} that 
similar words are used in similar contexts \citep{harris1954distributional}, thus assigning small cosine or euclidean distances between the
vector representations of ``great'' and ``awful''.
Our signed spectral normalized graph cut algorithm (henceforth, signed clustering) builds antonym relations into the vector space, while 
maintaining distributional similarity. Furthermore, another strength of K-clustering of signed graphs
is that it can be used collaboratively with other methods for augmenting semantic
meaning. Signed clustering leads to improved
clusters over spectral clustering of word embeddings, and has better coverage than thesaurus look-up.
This is because thesauri erroneously give equal weight to rare senses of word, such as ``absurd'' and its rarely used synonym ``rich''. 
Also, the overlap between thesauri is small, due to their manual creation.
\citet{lin1998automatic} found 0.178397 between-synonym set from Roget's Thesaurus and WordNet 1.5.
We also found similarly small overlap between all three thesauri tested.

We evaluated our clusters by comparing them to different
vector representations. In addition, we evaluated our clusters against SimLex-999.
Finally, we tested our method on the sentiment analysis task.
Overall, signed spectral clustering results are a very clean and 
elegant augmentation to current methods, and may have broad application to many fields.
Our main contributions are the novel method for signed clustering of signed graphs by \citet{gallier2015spectral},
the application of this method to create semantic word clusters which are agnostic to both vector space representations and thesauri, 
and finally, the systematic evaluation and creation of word clusters using thesauri.

 \subsection{Related Work}
 
Semantic word cluster and distributional thesauri have been well studied \citep{lin1998automatic,curran2004distributional}.
Recently there has been a lot of work on incorporating synonyms and antonyms into word embeddings.
Most recent models either attempt to make richer contexts, in order to find semantic similarity, \
or overlay thesaurus information in a supervised or semi-supervised manner.
\citet{tang2014learning} created sentiment-specific word embedding (SSWE), which
were trained for twitter sentiment. 
\citet{yih2012polarity} proposed 
polarity induced latent semantic analysis (PILSA) using thesauri,
which was extended by \citet{chang2013multi} to a multi-relational setting.
The Bayesian tensor factorization model (BPTF) was introduced in order to
combine multiple sources of information \citep{zhang2014word}.
\citet{faruqui2014retrofitting} used belief propagation to modify existing vector space representations.
The word embeddings on Thesauri and Distributional
information (WE-TD) model \citep{ono2015word} incorporated thesauri by 
altering the objective function for word embedding representations.
Similarly, \citet{marcobaroni2multitask} introduced multitask Lexical Contrast Model
which extended the word2vec Skip-gram method to optimize for both context as well as synonymy/antonym relations.
Our approach differs from the afore-mentioned methods in that
we created word clusters using the antonym relationships as negative links.
Similar to \citet{faruqui2014retrofitting} our signed clustering method uses existing vector representations to create word clusters.

To our knowledge, \citet{gallier2015spectral} is the first theoretical foundation of multiclass signed normalized cuts. 
\citet{hou2005bounds} used positive degrees of nodes in the degree matrix
of a signed graph with weights (-1, 0, 1), which was advanced by  
\citet{kolluri2004spectral,kunegis2010spectral} using absolute values of weights in the degree matrix. 
Although must-link and cannot-link soft spectral clustering \citep{rangapuramconstrained} both share similarities with our method, this similarity only applies to cases where cannot-link edges are present. Our method excludes a weight term of cannot-link, as well as the volume of cannot-link edges within the clusters.
Furthermore, our optimization method differs from that of must-link / cannot-link algorithms.
We developed a novel theory and algorithm that extends the 
clustering of \citet{shi2000normalized,yu2003multiclass} to the multi-class signed graph case \citep{gallier2015spectral}.


%% file: sections/theory.tex
\section{Signed Graph Cluster Estimation}

\subsection{Signed Normalized Cut}

Weighted graphs
for which the weight matrix is a symmetric matrix in which negative
and positive entries are allowed are called {\it signed graphs\/}.
Such graphs (with weights $(-1, 0, +1)$) were introduced as early as
1953 by  \cite{harary1953notion}, to model social relations involving disliking,
indifference, and liking.  The problem of clustering the nodes
of a signed graph arises naturally as a generalization of the
clustering problem for weighted graphs. Figure 1 shows a signed graph of word similarities with a thesaurus overlay.
\begin{figure}[ht]
  \label{fig:dc2}
  \includegraphics[width=\linewidth]{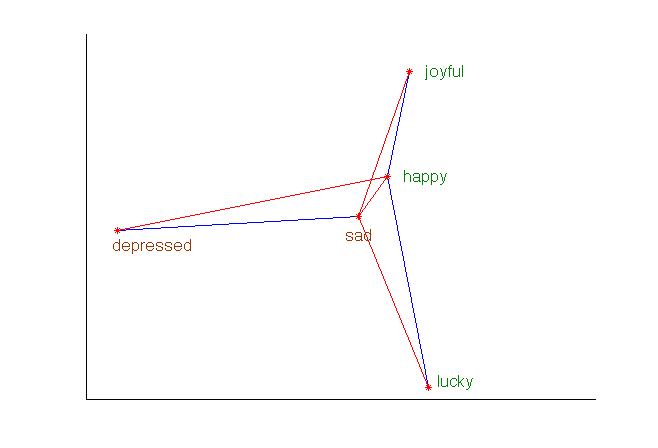}
  \caption{Signed graph of words using 
  a distance metric from the word embedding. The red edges represent the antonym relation while blue edges represent synonymy relations.}
\end{figure}
\citet{gallier2015spectral} extends normalized cuts signed graphs in order to incorporate antonym information into word clusters.

\theoremstyle{definition}
\begin{definition}
\label{graph-weighted}
A {\it  weighted graph\/} 
is a pair $G = (V, W)$, where 
$V = \{v_1,  \ldots, v_m\}$ is a set of
{\it nodes\/} or {\it vertices\/}, and $W$ is a symmetric matrix
called the {\it weight matrix\/}, such that $w_{i\, j} \geq 0$
for all $i, j \in \{1, \ldots, m\}$, 
and $w_{i\, i} = 0$ for $i = 1, \ldots, m$.
We say that a set $\{v_i, v_j\}$  is an edge iff
$w_{i\, j} > 0$. The corresponding (undirected) graph $(V, E)$
with $E = \{\{v_i, v_j\} \mid w_{i\, j} > 0\}$, 
is called the {\it underlying graph\/} of $G$.
\end{definition}

Given a signed graph $G = (V, W)$ (where $W$ is a symmetric matrix
with zero diagonal entries), the {\it underlying graph\/} of $G$ is
the graph with node set $V$ and set of (undirected) edges
$E = \{\{v_i, v_j\} \mid w_{i j} \not= 0\}$.

If $(V, W)$ is a signed graph, where $W$ is an $m\times m$ symmetric
matrix with zero diagonal entries and with the other entries
$w_{i j}\in \reals$ arbitrary, for any node $v_i \in V$, the {\it signed degree\/} of $v_i$ is defined as
\[
\overline{d}_i = \overline{d}(v_i) = \sum_{j = 1}^m |w_{i j}|,
\]
and the {\it signed degree matrix \/} $\overline{\Degsym}$ as
\[
\overline{\Degsym} = \mathrm{diag}(\overline{d}(v_1) , \ldots, \overline{d}(v_m)).
\]
For any subset $A$ of the set of nodes
$V$, let
\[
\mathrm{vol}(A) = \sum_{v_i\in A} \overline{d}_i = 
 \sum_{v_i\in A} \sum_{j = 1}^m |w_{i j}|.
\]
For any two subsets $A$ and $B$ of $V$, 
define $\mathrm{links}^+(A,B)$, 
$\mathrm{links}^-(A,B)$, and $\mathrm{cut}(A,\overline{A})$ by
\begin{align*}
\mathrm{links}^+(A, B) & = 
\sum_{\begin{subarray}{c}
v_i \in A, v_j\in B \\
w_{i j} > 0
\end{subarray}
}
w_{i j}  \\
\mathrm{links}^-(A,B) & = 
\sum_{\begin{subarray}{c}
v_i \in A, v_j\in B \\
w_{i j} < 0
\end{subarray}
}
- w_{i j} \\
\mathrm{cut}(A,\overline{A}) & = 
\sum_{\begin{subarray}{c}
v_i \in A, v_j\in \overline{A} \\
w_{i j} \not=  0
\end{subarray}
}
|w_{i j}| .
\end{align*}
Then, the {\it signed Laplacian\/} $\overline{L}$ is defined by
\[
\overline{L} = \overline{\Degsym} - W, 
\]
and its normalized version $\overline{L}_{\mathrm{sym}}$ by
\[
\overline{L}_{\mathrm{sym}} =  \overline{\Degsym}^{-1/2}\, \overline{L}\,
\overline{\Degsym}^{-1/2}
= I - \overline{\Degsym}^{-1/2} W \overline{\Degsym}^{-1/2}.  
\]
For a graph without isolated vertices, we have $\overline{d}(v_i) > 0$
for $i = 1, \ldots, m$, so $\overline{\Degsym}^{-1/2}$ is well defined.

\begin{proposition}
\label{Laplace1s}
For any  $m\times m$ symmetric matrix $W = (w_{i j})$, if we let $\overline{L} = \overline{\Degsym} - W$
where $\overline{\Degsym}$ is the signed degree matrix associated with $W$,
then  we have
\[
\transpos{x} \overline{L} x =
\frac{1}{2}\sum_{i, j = 1}^m |w_{i  j}| (x_i - \mathrm{sgn}(w_{i j}) x_j)^2
\quad\mathrm{for\ all}\> x\in \reals^m.
\]
Consequently, $\overline{L}$  is positive semidefinite.
\end{proposition}

Given a partition of $V$ into $K$
clusters $(A_1, \ldots, A_K)$, if we represent the $j$th block of
this partition by a vector $X^j$ such that
\[
X^j_i = 
\begin{cases}
a_j & \text{if $v_i \in A_j$} \\
0 &  \text{if $v_i \notin A_j$} ,
\end{cases}
\]
for some $a_j \not= 0$.

\begin{definition}
\label{sncut}
The {\it signed normalized cut\/}
$\mathrm{sNcut}(A_1, \ldots, A_K)$ of the
partition $(A_1, ..., A_K)$ is defined as
\[
\mathrm{sNcut}(A_1, \ldots, A_K) = \sum_{j = 1}^K
\frac{\mathrm{cut}(A_j, \overline{A_j}) + 
2 \mathrm{links}^-(A_j, A_j)}{\mathrm{vol}(A_j)}.
\] 
\end{definition}

Another formulation is
\[
\mathrm{sNcut}(A_1, \ldots, A_K) = 
\sum_{j = 1}^K \frac{\transpos{(X^j)} \overline{L} X^j} 
  {\transpos{(X^j)} \overline{\Degsym} X^j}.
\]
where $X$ is the $N\times K$ matrix whose $j$th column is $X^j$.

Observe that minimizing $\mathrm{sNcut}(A_1, \ldots, A_K)$ amounts to 
minimizing the number of positive and negative edges between clusters,
and also minimizing the number of negative edges within clusters.
This second minimization captures the intuition that nodes connected
by a negative edge should not be together  (they do not ``like''
each other; they should be far from each other).

\subsection{Optimization Problem}

We have our first formulation of $K$-way clustering
of a graph using normalized cuts, called problem PNC1 
(the notation PNCX  is used in  Yu \cite{yu2003multiclass}, Section 2.1):

If we let
\begin{align*}
\s{X}  = \Big\{[X^1\> \ldots \> X^K] \mid
X^j = a_j(x_1^j, \ldots, x_N^j) , \\
\>
x_i^j \in \{1, 0\},
 a_j\in \reals, \> X^j \not= 0
\Big\}
\end{align*}
our solution set is
\[
\s{K}  = \big\{
X  \in\s{X}  \mid \transpos{X}  \overline{\Degsym}\mathbf{1} = 0
\big\}.
\]

\medskip\noindent
{\bf $K$-way Clustering of a graph using Normalized Cut, Version 1: \\
Problem PNC1}


\begin{align*}
& \mathrm{minimize}     &  \sum_{j = 1}^K 
\frac{\transpos{(X^j)} \overline{L} X^j}{\transpos{(X^j)}\overline{\Degsym} X^j}& &  &  &\\
& \mathrm{subject\ to} & 
 \transpos{(X^i)} \overline{\Degsym} X^j = 0,  & &  &  &\\
& & \quad 1\leq i, j \leq K,\> 
i\not= j,  & &  X\in \s{X}. & &  & 
\end{align*}

An equivalent version of the optimization problem is

\medskip\noindent

{\bf Problem PNC2}
\begin{align*}
& \mathrm{minimize}     &  &  
\mathrm{tr}(\transpos{X} \overline{L} X)& &  &  &\\
& \mathrm{subject\ to} &  & 
\transpos{X} \overline{\Degsym} X = I, 
 & &  X\in \s{X}. & &
\end{align*}

The natural relaxation of problem PNC2 is to drop the condition
that $X\in \s{X}$, and we obtain the 

\medskip\noindent
{\bf Problem $(*_2)$}

\begin{align*}
& \mathrm{minimize}     &  &  
\mathrm{tr}(\transpos{X} \overline{L} X)& &  &  &\\
& \mathrm{subject\ to} &  & 
\transpos{X} \overline{\Degsym} X = I, 
 & &  & & 
\end{align*}

If $X$ is a solution to the relaxed problem, then $XQ$ is also a solution, where $Q\in\mathbf{O}(K)$.

If we make the change of variable $Y = \overline{\Degsym}^{1/2} X$ or equivalently
$X = \overline{\Degsym}^{-1/2} Y$.

However, since $\transpos{Y} Y = I$, we have
\[
Y^+ = \transpos{Y},
\]
so we get the equivalent problem

\medskip\noindent
{\bf Problem $(**_2)$}

\begin{align*}
& \mathrm{minimize}     &  &  
\mathrm{tr}(\transpos{Y}\overline{\Degsym}^{-1/2} \overline{L} \overline{\Degsym}^{-1/2} Y)& &  &  &\\
& \mathrm{subject\ to} &  & 
\transpos{Y} Y = I. 
 & &  & &  
\end{align*}

The minimum of  problem $(**_2)$
is achieved by any $K$ unit eigenvectors $(u_1, \ldots, u_K)$ associated with the smallest
eigenvalues
\[
0 = \nu_1\leq  \nu_2 \leq  \ldots \leq  \nu_K
\]
of $L_{\mathrm{sym}}$.

\subsection{Finding an Approximate Discrete Solution}

Given a solution $Z$ of problem $(*_2)$, 
we  look for pairs
$(X, Q)$ with $X\in \s{X}$ and  where $Q$ is a $K\times K$ matrix with
nonzero and pairwise orthogonal columns,
with $\norme{X}_F = \norme{Z}_F$,
that minimize
\[
\varphi(X, Q) = \norme{X - ZQ}_F.
\]
Here, $\norme{A}_F$ is the Frobenius norm of $A$.

This is a difficult nonlinear optimization problem 
involving two unknown matrices $X$ and $Q$. 
To simplify the problem,
we proceed by alternating steps during which we minimize 
$\varphi(X, Q) = \norme{X - ZQ}_F$ with respect to $X$ holding $Q$
fixed, and steps during which we minimize 
$\varphi(X, Q) = \norme{X - ZQ}_F$ with respect to $Q$ holding $X$
fixed. 

This second step in which $X$ is held fixed has been studied, but it
is still a hard problem for which no closed--form solution is known.
Consequently, we further simplify the problem.
Since $Q$ is of the form $Q = R\Lambda$ where
$R\in \mathbf{O}(K)$ and $\Lambda$ is a diagonal invertible matrix,
we minimize $\norme{X - ZR\Lambda}_F$ in two stages.
\begin{enumerate}
\item
We set $\Lambda = I$ and find $R\in \mathbf{O}(K)$
that minimizes  $\norme{X - ZR}_F$.
\item
Given $X$, $Z$, and $R$,  find a 
diagonal invertible matrix $\Lambda$ that
minimizes  $\norme{X - ZR\Lambda}_F$.
\end{enumerate}

The matrix $R\Lambda$ is not a minimizer of
$\norme{X - ZR\Lambda}_F$ in general, but it is an improvement
on $R$ alone, and both stages can be solved quite easily.

In stage 1, the matrix $Q=R$ is orthogonal, so $Q\transpos{Q} = I$, and
since $Z$ and $X$ are given, 
the problem reduces to minimizing
$ - 2\mathrm{tr}(\transpos{Q}\transpos{Z}X)$; that is,
maximizing   $\mathrm{tr}(\transpos{Q}\transpos{Z}X)$.


%% file: sections/metrics.tex
\section{Metrics}

The evaluation of clusters is non-trivial to generalize. We used both intrinsic and extrinsic methods of evaluation.
Intrinsic evaluation is two fold where we only examine cluster entropy, purity, number of disconnected components and 
number of negative edges. We also compare multiple word embeddings and thesauri to show stability of our method.
The second intrinsic measure is using a gold standard. We chose a gold standard 
designed for the task of capturing word similarity. Our metric for evaluation is a detailed
accuracy and recall. 
For extrinsic evaluation, we use our clusters to
identify polarity and apply this to the task. 

\subsection{Similarity Metric and Edge Weight}
For clustering there are several choices to make. The first choice being the similarity metric.
In this paper we chose the heat kernel based off of Euclidean distance between word vector
representations. We define the distance between two words $w_i$ and $w_j$ as 
$dist(w_i, w_j) = \norme{w_i - w_j}$.
In the paper by \citet{belkin2003laplacian}, the authors show that the 
heat kernel where
\[
W_{ij} = 
\begin{cases}
0 & \text{ if } e^{-\frac{dist(w_i, w_j)^2}{\sigma}} < thresh \\
e^{-\frac{dist(w_i, w_j)^2}{\sigma}} & \text{ otherwise}
\end{cases}.
\]


The next choice of how to combine the word embeddings with the thesauri in order to make a signed 
graph also has hyperparameters. We can represent the thesaurus as a matrix where 
\[
T_{ij} = 
\begin{cases}
1 & \text{ if words $i$ and $j$ are synonyms} \\
-1 & \text{ if words $i$ and $j$ are antonyms} \\
0 & \text{ otherwise} \\
\end{cases}.
\]
Another alternative is to only look at the antonym information, so
\[
T^{ant}_{ij} = 
\begin{cases}
-1 & \text{ if words $i$ and $j$ are antonyms} \\
0 & \text{ otherwise} \\
\end{cases}.
\]

We can write the signed graph as $\hat{W_{ij}} = \beta T_{ij} W_{ij}$ or in matrix form
$\hat{W} =\beta T \odot W$  where $\odot$ computes Hadamard product (element-wise multiplication); however,
the graph will only contain the overlapping vocabulary. 
In order to solve this problem we use $\hat{W} = \gamma W + \beta^{ant} T^{ant} \odot W  + \beta T \odot W$.

\subsection{Evaluation Metrics}
%
%

It is important to note that this metric does not require a gold standard.
Obviously we want this number to be as small as possible.

As we used thesaurus information for two other novel metrics which are the number of
negative edges (NNE) in the clusters, and the number of disconnected components (NDC) in the cluster where we only use synonym edges. 
\begin{align*}
NDC &= \sum_{r=1}^{k}{\sum_{i=1}^{C}{(n_r^i)}}
\end{align*}

The NDC has the disadvantage of thesaurus coverage. 
Figure 2 shows a graphical representation of the number of disconnected components and negative edges.

\begin{figure}[ht]
  \label{fig:dc2}
  

  \begin{subfigure}
  \par
    \label{fig:figure2}
    \includegraphics[width=\linewidth]{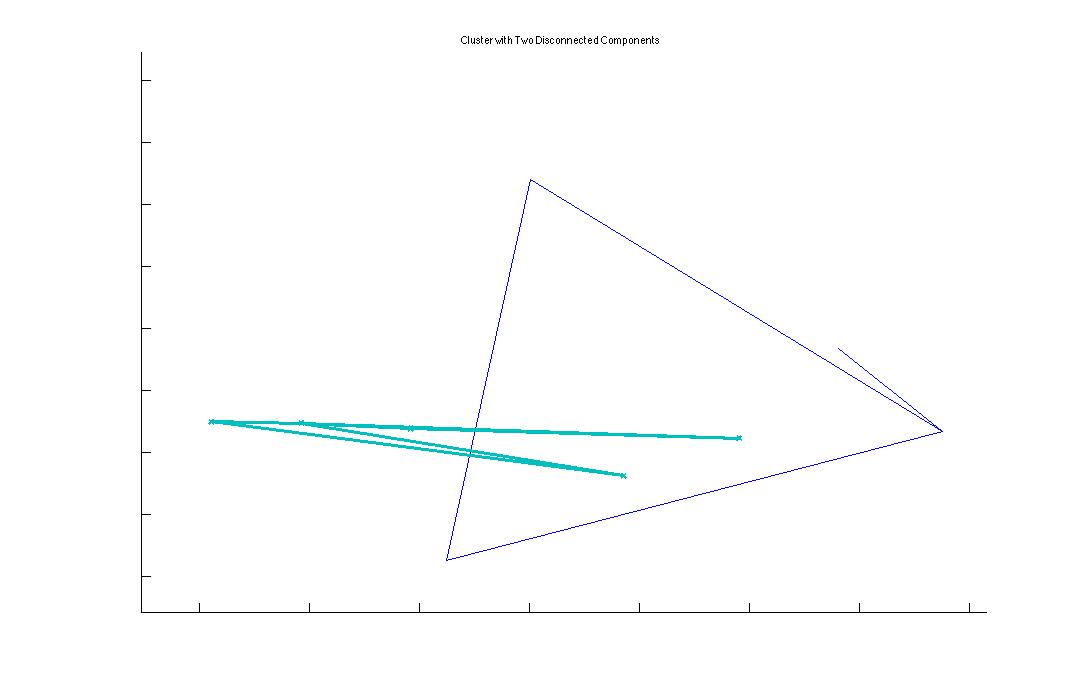}
    {\small {\it Figure 2.1.} Cluster with two disconnected components. 
    All edges represent synonymy relations. The edge colors are only meant to highlight the different components.}
  \end{subfigure}
  \begin{subfigure}
  \par
    \label{fig:ant1}
    \includegraphics[width=\linewidth]{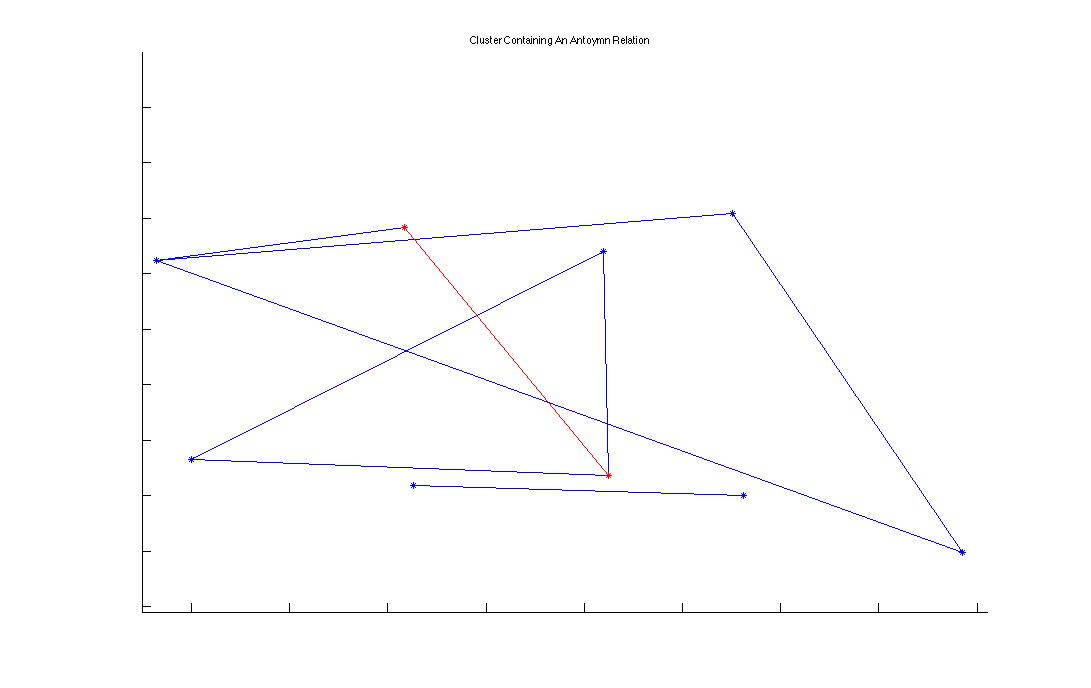}
    {\small {\it Figure 2.1.} Cluster with one antonym relation. 
    The red edge represents the antonym relation. Blue edges represent synonymy relations.}
  \end{subfigure}
  \caption{Disconnected component and number of antonym evaluations.}
\end{figure}

Next we evaluate our clusters using an external gold standard. 
Cluster purity and entropy \cite{zhao2001criterion} is defined as, 
\begin{align*}
Purity &= \sum_{r=1}^{k}{\frac{1}{n}max_i(n_r^i)} \\
Entropy &= \sum_{r=1}^{k}{\frac{n_r}{n}\left({-\frac{1}{\log q}\sum_{i=1}^{q}{\frac{n_r^i}{n_r}\log \frac{n_r^i}{n_r}}}\right)}
\end{align*}

where $q$ is the number of classes, $k$ the number of clusters, $n_r$ is the size of cluster $r$, and $n_r^i$
number of data points in class $i$ clustered in cluster $r$.
The purity and entropy measures improve (increased purity, decreased entropy) monotonically with the number of clusters.

%% file: sections/results.tex
\section{Empirical Results}

In this section we begin with intrinsic analysis of the resulting clusters. 
We then compare empirical clusters with 
SimLex-999 as a gold standard for semantic word similarity.
Finally, we evaluate our metric using the sentiment prediction task.
Our synonym clusters are well suited for
this task, as including antonyms in clusters results in incorrect predictions.

\subsection{Simulated Data}
In order to evaluate our signed graph clustering method, we first focused on intrinsic measures of cluster quality.
 Figure 3.2  demonstrates that the number of negative edges within a cluster is minimized using our clustering algorithm on simulated data. However, as the number of clusters becomes large, the number of disconnected components, which includes clusters of size one, increases. For our further empirical analysis, we used both the number of disconnected components as well as the number of antonyms within clusters in order to set the cluster size.

\begin{figure}[ht]
  \label{fig:dcnemetric}
  \begin{subfigure}
  \par
    \label{fig:simgraph}
    \includegraphics[width=\linewidth]{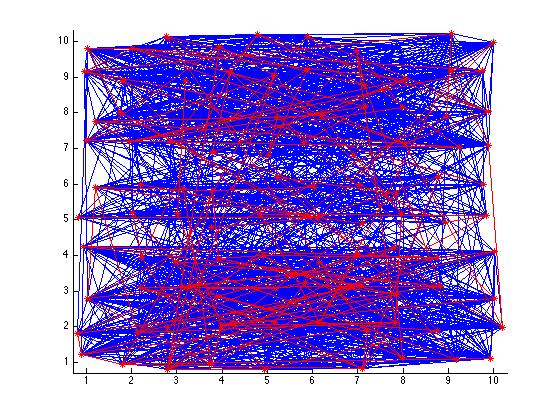}
    {\small {\it Figure 3.1.} Simulated signed graph \ \ \ \ \ \ \ \ \ \ \ \ \ \ \ \ \ \ \ \ \ \ \ \ \ \ \ }
  \end{subfigure}
  \begin{subfigure}
  \par
    \label{fig:ant1}
    \includegraphics[width=\linewidth]{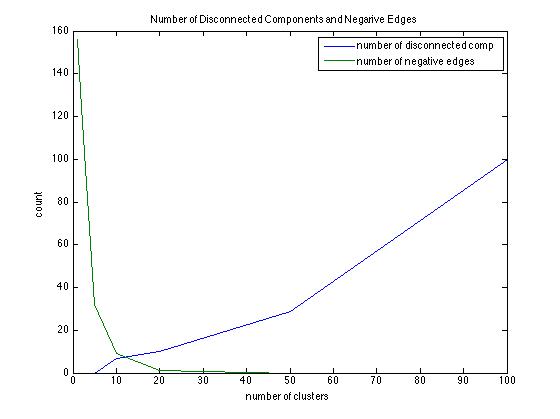}
    {\small {\it Figure 3.2.} This is a plot of the relationship between the number of disconnected components and negative edges within the clusters.}
  \end{subfigure}
  \caption{Graph of Disconnected Component and Negative Edge Relations}
\end{figure}

\subsection{Data}
\subsubsection{Word Embeddings}
For comparison, we used four different word embedding methods: Skip-gram vectors (word2vec) \citep{mikolov2013efficient}, Global vectors (GloVe) \citep{pennington2014glove}, Eigenwords \citep{dhillon2015eigenwords}, and Global Context (GloCon) \citep{huang2012improving} vector representation. 
We used word2vec 300 dimensional embeddings which were trained using word2vec code on several billion words of English comprising the entirety of Gigaword and the English discussion forum data gathered as part of BOLT. A minimal tokenization was performed based on CMU's twoknenize\footnote{\url{https://github.com/brendano/ark-tweet-nlp}}. 
For GloVe we used pretrained 200 dimensional vector embeddings\footnote{\url{http://nlp.stanford.edu/projects/GloVe/}} trained using Wikipedia 2014 + Gigaword 5 (6B tokens). 
Eigenwords were trained on English Gigaword with no lowercasing or cleaning.
Finally, we used 50 dimensional vector representations from \citet{huang2012improving}, which
used the April 2010 snapshot of the Wikipedia corpus \cite{lin1998automatic,shaoul2010westbury},
with a total of about 2 million articles and 990 million tokens.

\subsubsection{Thesauri}
Several thesauri were used, in order to test robustness (including Roget's Thesaurus,
the Microsoft Word English (MS Word) thesaurus from \citet{samsonovic2010principal} and WordNet 3.0) \citep{miller1995wordnet}.
\citet{jarmasz2004roget,hale1998comparison} have shown that Roget's thesaurus has better semantic similarity than WordNet.
This is consistent with our results using a larger dataset of SimLex-999. 

We chose a subset of 5108 words for the training dataset, which had high overlap between various sources. Changes to the training dataset had minimal effects on the optimal parameters. 
Within the training dataset, each of the thesauri had roughly 3700 antonym pairs, and combined they had 6680. However, the number of distinct connected components varied, with Roget's Thesaurus having the least (629), and MS Word Thesaurus (1162) and WordNet (2449) having the most. These ratios were consistent across the full dataset.

\subsection{Cluster Evaluation}

One of our main goals was to go beyond qualitative analysis into quantitative measures of 
synonym clusters and word similarity. 
In Table \ref{tab:qualclusters}, we show the 4 most-associated words with ``accept'', ``negative'' and ``unlike''.

\begin{table*}[!tbh]
  \centering
  \small
  \begin{tabular}{|l||l|l|l|l|l|l|l|}
  \hline 
{\bf Ref word} & {\bf Roget} & {\bf WordNet} & {\bf MS Word} & {\bf W2V} & {\bf GloDoc} & {\bf EW} & {\bf Glove} \\ \hline
%
%
accept 
& adopt & agree & take & accepts & seek & approve & agree \\ 
& accept your fate & get & swallow & reject & consider & declare & reject \\
& be fooled by & fancy & consent & agree & know & endorse & willin \\ 
& acquiesce & hold & assume & accepting & ask & reconsider & refuse \\
negative
& not advantageous & unfavorable & severe & positive & reverse & unfavorable & positive \\
& pejorative & denial & hard & adverse & obvious & positive & impact \\ 
& pessimistic & resisting & wasteful & Negative & calculation & dire & suggesting \\ 
& no & pessimistic & charged & negatively & cumulative & worrisome & result \\ 
\hline
unlike 
&  {\bf no synonyms} & incongruous & different & Unlike & whereas & Unlike & instance \\
& & unequal & dissimilar & Like & true & Like & though \\ 
& & separate &  & even & though & Whereas & whereas \\ 
& & hostile &  & But & bit & whereas & likewise \\ 
\hline

 \end{tabular}
   \caption{Qualitative comparison of clusters.}
  \label{tab:qualclusters}
\end{table*}

\subsubsection{Cluster Similarity and Hyperparameter Optimization}

For a similarity metric between any two words, we use the heat kernel of Euclidean distance, so $sim(w_i, w_j) = e^{-\frac{\norme{w_i-w_j}^2}{\sigma}}$.
The thesaurus matrix entry $T_{ij}$ has a weight of 1 if words $i$ and $j$ are synonyms, -1  if words $i$ and $j$ are antonyms, and 0 otherwise. Thus the weight matrix entries $W_{ij} = T_{ij}e^{-\frac{\norme{w_i-w_j}^2}{\sigma}}$.

\begin{table*}[!tbh]
  \centering
  \small
  \begin{tabular}{|l||p{1cm}|c|c|c|c|c|}
  \hline 
\multicolumn{1}{|c||}{\bf Method} & 
{\bf $\sigma$} & 
{\bf thresh} & 
{\bf \# Clusters} & 
{\bf Error  $\downarrow$ } & 
{\bf Purity $\uparrow$} &
{\bf Entropy $\downarrow$}   \\
                 &       &      &     & $\frac{(NNE+NDC)}{|V|}$ & & \\ \hline 
Word2Vec         & 0.2   & 0.04 & 750  & 0.716 & 0.88 & 0.14 \\ \hline
Word2Vec + Roget & 0.7   & 0.04 & 750  & 0.033 & 0.94 & 0.09 \\ \hline 
Eigenword        & 2.0   & 0.07 & 200  & 0.655 & 0.84 & 0.25 \\ \hline
Eigenword + MSW  & 1.0   & 0.08 & 200  & 0.042 & 0.95 & 0.01 \\ \hline 
GloCon            & 3.0   & 0.09 & 100  & 0.691 & 0.98 & 0.03 \\ \hline
GloCon + Roget    & 0.9   & 0.06 & 750  & 0.048 & 0.94 & 0.02 \\ \hline 
Glove            & 9.0   & 0.09 & 200  & 0.657 & 0.72 & 0.33 \\ \hline
Glove + Roget    & 11.0  & 0.01 & 1000 & 0.070 & 0.91 & 0.10 \\ \hline
\end{tabular}
   \caption{Clustering evaluation after parameter optimization minimizing error using grid search.}
  \label{tab:optimalparams}
\end{table*}

%
Table \ref{tab:optimalparams} shows results from the grid search of hyperparameter optimization.
Here we show that Eigenword + MSW outperforms Eigenword + Roget, which is in contrast 
with the other word embeddings where the combination with Roget performs better. 

As a baseline, we created clusters using K-means where the number of K clusters was set to 750. 
All K-means clusters have a statistically significant difference in the number of antonym pairs relative to
random assignment of labels. When compared with the MS Word thesaurus, Word2Vec, Eigenword, GloCon, and GloVe word embeddings had a total of 286, 235, 235, 220 negative edges, respectively. The results are similar with the other thesauri. This shows that there are a significant number of antonyms pairs in the K-means clusters derived from the word embeddings. By optimizing the hyperparameters using normalized cuts without thesauri information, we found a significant decrease in the number of negative edges, which was indistinguishable from random assignment and corresponded to a roughly ninety percent decrease across clusters. When analyzed using an out of sample thesaurus and 27081 words, the number of antonym clusters decreased to under 5 for all word embeddings, with the addition of antonym relationship information. 

If we examined the number of distinct connected components within the different word clusters, we
observed that when K-means were used, the number of disconnected components were statistically significant from random labelling. This suggests that the word embeddings capture synonym relationships. By optimizing the hyperparameters we found roughly a 10 percent decrease in distinct connected components using normalized cuts. When we added the signed antonym relationships using our signed clustering algorithm, on average we found a thirty-nine percent decrease over the K-means clusters. Again, this shows that the hyperparameter optimization is highly effective.




    
    
    


\subsubsection{Evaluation Using Gold Standard}

SimLex-999 is a gold standard resource for semantic similarity, not relatedness, based on ratings by human annotators. 
The differentiation between relatedness and similarity was a problem with previous datasets such as WordSim-353.
\citet{hill2014simlex} has a further comparison of SimLex-999 to previous datasets. 
Table \ref{tab:simlex} shows the difference between SimLex-999 and WordSim-353.
\begin{table*}[!tbh]
  \centering
  \small
  \begin{tabular}{|c||c|c|}
  \hline 
{\bf Pair} & 	{\bf Simlex-999 rating} & {\bf	WordSim-353 rating} \\ \hline
coast - shore &	9.00 &	9.10 \\ \hline
clothes - closet & 	1.96 & 8.00 \\ \hline
\end{tabular}
   \caption{Comparison between SimLex-999 and WordSim-353.
   This is from \url{http://www.cl.cam.ac.uk/~fh295/simlex.html}}.
  \label{tab:simlex}
\end{table*}
SimLex-999 comprises of multiple parts-of-speech with 666 Noun-Noun pairs, 222 Verb-Verb pairs and 111 Adjective-Adjective pairs. 
In a perfect setting, all word pairs rated highly similar by human annotators would be in the same cluster, and all words which were rated dissimilar would be in different clusters. 
Since our clustering algorithm  produced sets of words, we used this evaluation instead of the more commonly-reported correlations.
%



\begin{table}[!tbh]
  \centering
  \small
  \begin{tabular}{|l||c|c|}
  \hline 
\multicolumn{1}{|c||}{\bf Method} & {\bf Accuracy} & {\bf Coverage} \\ \hline
MS Thes Lookup & 0.70 & 0.57 \\ \hline
Roget Thes Lookup & 0.63 & 0.99  \\ \hline
WordNet Thes Lookup & 0.43 & 1.00 \\ \hline
Combined Thes Lookup & 0.90 & 1.00 \\ \hline
Word2Vec & 0.36 & 1.00 \\ \hline
Word2Vec+CombThes & 0.67 & 1.00 \\ \hline
Eigenwords & 0.23 & 1.00  \\ \hline
Eigenwords+CombThes & 0.12 & 1.00  \\ \hline
GloCon & 0.07 & 1.00  \\ \hline
GloCon+CombThes & 0.05 & 1.00  \\ \hline
GloVe & 0.33 & 1.00  \\ \hline
GloVe+CombThes & 0.58 & 1.00 \\ \hline
Thes Lookup+W2V+CombThes & 0.96 & 1.00 \\ \hline
\end{tabular}
   \caption{Clustering evaluation using SimLex-999 with 120 word pairs having similarity score over 8.}
  \label{tab:simlexeval}
\end{table}

In Table \ref{tab:simlexeval} we show the results of the evaluation with SimLex-999. 
Accuracy increased for all of the clustering methods aside from Eigenwords+CombThes. However, we achieved better results when we exclusively used the MS Word thesaurus. 
Combining thesaurus lookup and word2vec+CombThes clusters yielded an accuracy of 0.96. 

\subsubsection{Sentiment Analysis}
We used the \citet{socher2013recursive}
sentiment treebank \footnote{\url{http://nlp.stanford.edu/sentiment/treebank.html}} with coarse grained labels on phrases and
sentences from movie review excerpts.
The treebank is split into training (6920) , development (872), and test (1821)
datasets. 
We trained an $l_2$-norm regularized logistic regression \citep{friedman2001elements} using our word clusters 
in order to predict the coarse-grained sentiment at the sentence level. 
We compared our model against existing models: Naive
Bayes with bag of words (NB), 
sentence word embedding averages (VecAvg),
retrofitted sentence word embeddings (RVecAvg) \citep{faruqui2014retrofitting},
simple recurrent neural network (RNN),
recurrent neural tensor network (RNTN) \citep{socher2013recursive}, 
and the state-of-the art Convolutional neural network (CNN) \citep{kim2014convolutional}.
Table \ref{tab:sentanalysis} shows that although our model
does not out-perform the state-of-the-art,
signed clustering performs better than comparable models, including the recurrent neural network,
which has access to more information. 


\begin{table}[!tbh]
  \centering
  \small
  \begin{tabular}{|l||c|c|}
  \hline 
\multicolumn{1}{|c||}{\bf Model} & {\bf Accuracy} \\ \hline
NB \citep{socher2013recursive} & 0.818 \\ \hline 
VecAvg (W2V, GV, GC)  & 0.812, 0.796, 0.678  \\ 
\citep{faruqui2014retrofitting} & \\ \hline
RVecAvg (W2V, GV, GC)  & 0.821, 0.822, 0.689  \\  
\citep{faruqui2014retrofitting} & \\ \hline
RNN, RNTN \citep{socher2013recursive} & 0.824, 0.854 \\ \hline
CNN \citep{le2015compositional} & 0.881 \\ \hline 
\hline
SC W2V & 0.836 \\ \hline
SC GV & 0.819 \\ \hline
SC GC & 0.572 \\ \hline
SC EW & 0.820 \\ \hline
\end{tabular}
   \caption{Sentiment analysis accuracy for binary predictions of signed clustering algorithm (SC) versus other models.}
  \label{tab:sentanalysis}
\end{table}




%% file: sections/conclusion.tex
\section{Conclusion}

We developed a novel theory for signed normalized cuts
as well as an algorithm for finding the discrete solution.
We showed that we can find superior synonym clusters which
do not require new word embeddings, but simply overlay thesaurus information.
The clusters are general and can be used with several out of the box word embeddings. 
By accounting for antonym relationships, our algorithm greatly outperforms simple normalized cuts, even with Huang's word embeddings
, which are designed to capture semantic relations.
Finally, we examined our clustering method on the sentiment analysis task from \citet{socher2013recursive} sentiment treebank dataset 
and showed
improved performance versus comparable models.

This method could be applied to a broad range of NLP tasks, such as prediction of social group clustering, identification of personal versus non-personal verbs, and analysis of clusters which capture positive, negative, and objective emotional content. It could also be used to explore multi-view relationships, such as aligning synonym clusters across multiple languages. Another possibility is to use thesauri and word vector representations together with word sense disambiguation
to generate synonym clusters for multiple senses of words.
Finally, our signed clustering could be extended to evolutionary signed clustering.